\pdfoutput=1

\documentclass[11pt]{article}
\usepackage[table,xcdraw]{xcolor}

\usepackage[]{acl}

\usepackage{times}
\usepackage{latexsym}

\usepackage[T1]{fontenc}

\usepackage[utf8]{inputenc}

\usepackage{microtype}

\usepackage{tikz}
\usepackage{comment}
\usepackage{booktabs}
\usepackage{amsmath,amssymb} 
\usepackage{color}
\usepackage[capitalize]{cleveref}
\usepackage{algpseudocode}
\usepackage{multirow}
\usepackage{multicol}
\usepackage{makecell}
\usepackage{enumitem}
\usepackage{float}
\usepackage{wrapfig}
\usepackage{bbding}
\usepackage[normalem]{ulem}
\usepackage{balance}
\usepackage{adjustbox}
\useunder{\uline}{\ul}{}
\usepackage{nameref}
\usepackage{graphicx}
\usepackage{subcaption}

\title{Movie101v2: Improved Movie Narration Benchmark}

\author{Zihao Yue\thanks{Equal contribution. }, Yepeng Zhang$^*$, Ziheng Wang, Qin Jin \\
Renmin University of China\\
\small{\texttt{\{yzihao, yepzhang, zihengwang, qjin\}@ruc.edu.cn}}\\
\textcolor[rgb]{0.285,0.258,0.891}{\url{https://movie101-dataset.github.io}}}

\begin{document}
\maketitle

\begin{abstract}
Automatic movie narration aims to generate video-aligned plot descriptions to assist visually impaired audiences. 
Unlike standard video captioning, it involves not only describing key visual details but also inferring plots that unfold across multiple movie shots, presenting distinct and complex challenges. 
To advance this field, we introduce \textbf{Movie101v2}, a large-scale, bilingual dataset with enhanced data quality specifically designed for movie narration. 
Revisiting the task, we propose breaking down the ultimate goal of automatic movie narration into three progressive stages, offering a clear roadmap with corresponding evaluation metrics.
Based on our new benchmark, we baseline a range of large vision-language models, including GPT-4V, and conduct an in-depth analysis of the challenges in narration generation. 
Our findings highlight that achieving applicable movie narration generation is a fascinating goal that requires significant research.

\end{abstract}
\section{Introduction}

Audio Description (AD) is a crucial technology that enables visually impaired individuals to enjoy movies by integrating real-time voice narrations into the movie soundtrack, describing the movie's visual content. Unlike video captions, movie narrations must briefly summarize the ongoing content during pauses in character dialogue, providing accurate descriptions of visual facts (e.g., scenes, characters, and events) and key plot points to help the audience stay engaged. However, creating movie narrations usually requires extensive manual effort from trained professionals, making it impractical to cover the vast number of movies and TV shows online. Therefore, researchers have begun exploring automatic movie narration generation, advancing research from \textbf{data}~\cite{m-vad,mad}, \textbf{task}~\cite{lsmdc,autoad}, and \textbf{method}~\cite{autoad-ii,movie101} perspectives, but there is still a long way to go before realizing the ultimate goal of automatically generating high-quality and applicable movie narrations.

\begin{figure*}[t]
    \begin{center}
        \includegraphics[width=0.95\linewidth]{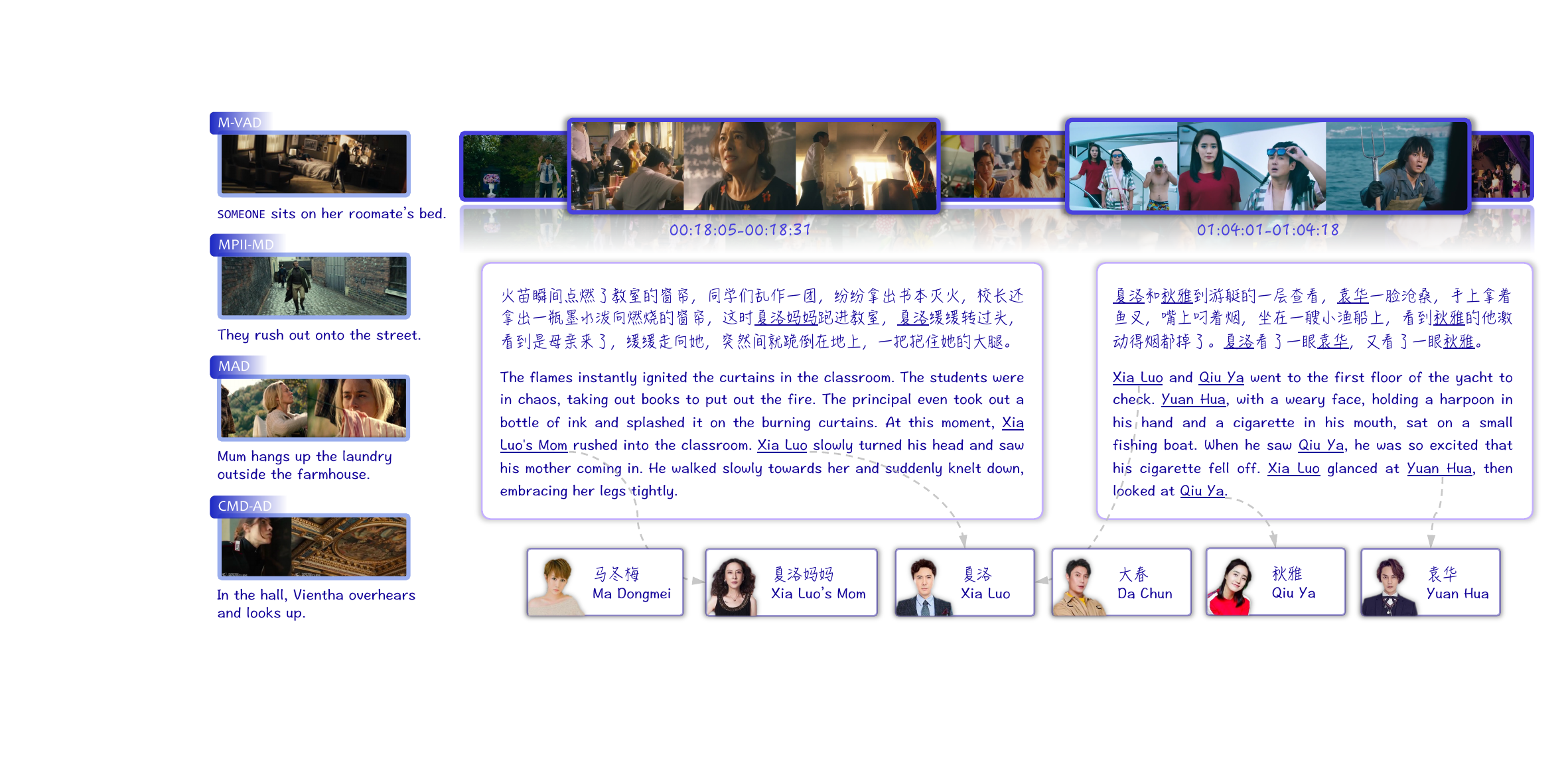}
    \end{center}
    \vspace{-10pt}
\caption{Examples from other datasets (left) and Movie101v2 (right) where cases are from \textit{Goodbye Mr. Loser}.}
\label{fig:data-main}
\vspace{-8pt}
\end{figure*}

\noindent \textbf{From the data perspective}, proper benchmark datasets play a fundamental role in the development of movie understanding and narrating. Early efforts primarily focus on basic video-to-text descriptions with simplified narration data~\cite{lsmdc,mad}. For example, LSMDC~\cite{lsmdc} replaces character names in narrations with "someone", reducing the movie narration task to a general video captioning task. AutoAD~\cite{autoad} improves on this by reinstating character names from the raw data of MAD~\cite{mad}, while AutoAD~II~\cite{autoad-ii} goes further by introducing a character bank to support generating narrations with character names. However, the movie clips designated for narration in these datasets are typically short (e.g., on average 6.2 seconds in M-VAD~\cite{m-vad} and 4.1 seconds in MAD), as shown in \cref{fig:data-main}, which limits the ability of models to generate coherent narrations for longer clips based on complex visual changes and plot sequences. 

The recently proposed dataset, Movie101~\cite{movie101}, addresses some of these limitations by providing longer video narration paragraphs and rich movie metadata, such as character information. However, our analysis identifies several drawbacks. First, Movie101 consists of only 101 movies with 14,000 video-narration pairs, which is smaller in scale compared to other available datasets. Second, it includes only Chinese narrations, limiting its usefulness for non-Chinese speakers and hindering the application of many advanced English-based models for movie understanding. Third, the automatically crawled metadata contains errors, such as missing characters in cast lists and inconsistent character names in narrations. To address these issues and to facilitate automatic movie narration from a data perspective, we improve Movie101 in terms of both \textit{scale} and \textit{quality}. By collaborating with various expert models, we manage to avoid the previously used heavy crowd-sourcing and increase the dataset cost-effectively to a scale of 203 movies and 46,000 bilingual (Chinese and English) video-narration pairs, which we name \textbf{Movie101v2}. 

\noindent \textbf{From the task perspective}, the development of automatic movie narration has evolved over time. AutoAD highlights the dependence of movie narration on context, incorporating a preceding narration and movie subtitles as additional task inputs. Movie101 and AutoAD~II address practical deployment issues of \textit{where} to insert narrations, by narrating between character dialogues or incorporating a timestamp prediction task. While these task definitions strive for more applicable narrations in real-world settings, our closer inspection of the Movie101v2 data reveals that achieving fully deployable movie narrations remains an ambitious goal.  

On the one hand, generating applicable narrations requires handling complex inputs like extensive plot histories and character dialogues, which are beyond the current capabilities of most models. On the other hand, our initial experimental findings suggest that even understanding the basic visual facts and movie plots within \textit{individual} movie clips is a fundamental but unsolved challenge. Therefore, the ambitious goal of achieving applicable movie narrations will require incremental progress to achieve step by step. To address this, we suggest breaking down the long-term goal into three progressively challenging stages: basic visual facts description (L1), plot narration (L2), and applicable AD text generation (L3). 
Given the current technological limitations, we prioritize achieving the L1 and L2 goals, focusing on movie understanding within individual clips.

To align with these staged goals, 
the evaluation framework should also evolve to provide more targeted feedback for model improvement. Existing works typically assess the linguistic or semantic matching between generated narrations and reference ones (ground truth)~\cite{lsmdc,autoad,autoad-ii,movie101}. 
However, since reference narrations are produced by human experts with access to extensive contextual information (e.g., prior plot developments and character histories), directly comparing them to model outputs, which are generated based solely on the given video clip, may not be a fair evaluation. 
To address this, we propose a new evaluation framework that leverages Large Language Models (LLMs) to separately assess narrative quality from the L1 and L2 perspectives. This helps avoid comparing against out-of-context information, and offers a clearer guidance for model development and improvement.

\noindent  \textbf{From the method perspective}, we explore practical solutions for movie narration generation. We build baselines based on several state-of-the-art Large Vision-Language Models (LVLMs), including both open-source models and GPT-4V~\cite{chatgpt}, on Movie101v2 in both Chinese and English. This provides a comprehensive evaluation of their performance in movie narration generation. 
We also carry out detailed analytical experiments to identify the challenges and difficulties that current models face, 
both in terms of visual perception and text generation. These findings aim to provide practical insights and inspire future research into improving automatic movie narration generation.

In summary, in our pursuit of advancing movie narration generation, large-scale and high-quality data are the cornerstone, clearly defined tasks and evaluations propel progress, and benchmarking and analysis of advanced models share important findings. Through the development of Movie101v2 in this work, our contributions in data, task design, evaluation framework, and experimental insights aim to shed some light on further research and progress in movie narration generation.
\section{Data: Movie101v2}

\begin{table}[t]
\caption{\label{tab:data-stat}
Dataset statistics. $L_{\mathrm{v/t}}$: average video duration (sec.) or text length (en. words or zh. characters); $N_{\mathrm{char}}$: character count; Gray: datasets with short movie clips.
}
\vspace{-4pt}
\centering
\small
\scalebox{0.97}{
\begin{tabular}{@{}l|cc|ccc@{}}
\toprule
Dataset & \# movie & \# clip & $L_{\mathrm{v}}$ & $L_{\mathrm{t}}$ & $N_{\mathrm{char}}$ \\ \midrule
\cellcolor[HTML]{EFEFEF}M-VAD & 92 & 49K & 6.2 & 9.1 & - \\
\cellcolor[HTML]{EFEFEF}MPII-MD & 94 & 68K & 3.9 & 9.6 & - \\
\cellcolor[HTML]{EFEFEF}MAD & 650 & 385K & 4.1 & 12.7 & - \\
\cellcolor[HTML]{EFEFEF}CMD-AD & 1,432 & 101K & - & - & - \\
Movie101 (zh) & 101 & 14K & 20.4 & 80.7 & 2.0 \\ \midrule
Movie101v2-zh & 203 & 46K & 12.8 & 60.0 & 1.9 \\
Movie101v2-en & 203 & 46K & 12.8 & 39.1 & 1.9 \\ \bottomrule
\end{tabular}}
\end{table}

The original Movie101 dataset contains 101 movies from the barrier-free channel of Xigua Video\footnote{\url{https://www.ixigua.com/channel/barrier_free}}, where movies are reproduced with video-aligned narration speeches. The narration texts are obtained through ASR transcription and refined with crowd-sourcing. In this section, we improve upon Movie101 in both data scale and quality.

\subsection{Scaling Up}

\label{sec:2.1.1}

Following Movie101, we collect all the newly available movies with narrations from Xigua Video (102 new movies in total), obtaining narration data through the following automatic process.

\noindent \textbf{Speech Transcription.} We transcribe movie audio into text by Whisper~\cite{whisper}. The raw ASR output includes both narrations and character dialogues, and often contains transcription errors.

\noindent \textbf{Text Refinement.} To remove character dialogues from the ASR outputs, we first use OCR~\cite{paddleocr} to detect movie subtitles, thereby identifying the instances where dialogues occur and subsequently removing them. This process concurrently produces subtitle data equipped with precise timestamps. Next, we use  GPT-4 to identify and remove any remaining dialogues. Upon filtering out dialogues, we leverage GPT-3.5-turbo to correct textual errors within the narrations, including typos, wrong punctuation, and nonsensical phrases.

\noindent \textbf{Clip Merging.} To support comprehension of coherent plots conveyed in consecutive movie clips, we merge adjacent narration segments into paragraphs. Employing a heuristic approach, we utilize a dynamically adjusted threshold to determine if two adjacent segments should be merged, thereby ensuring effective merging near clips while preventing the creation of excessively long paragraphs.

\noindent \textbf{Translation.} Following data refinement, we translate Chinese narrations into English using GPT-3.5-turbo. To ensure accurate translation of character names within narrations, we manually construct an English version of the movie casts, serving as references for the LLM.

\subsection{Quality Enhancement}
\label{sec:2.1.2}

\noindent \textbf{Character Name Refinement}. We collect movie metadata, including basic movie information and cast details, from Xigua Video. However, our manual review reveals two significant issues not identified in Movie101: (1) The cast list for some movies is incomplete, with some or all character names missing. (2) The character names in narrations often do not match their official cast names. Additionally, the same character may be narrated by different names throughout a movie, further limiting the connection between narrated characters and external cast knowledge. 
To address these issues, we refine character names in both the movie casts and narrations for Movie101 and our newly collected data. First, we complete and correct names in the cast lists through human annotation. Then, we use GPT-3.5-turbo to automatically update the narrations, ensuring the character names in the narrations precisely match those in the movie casts.


\noindent \textbf{Quality Control.} While our data construction leverages LLMs to minimize labor costs, it is crucial to ensure and monitor the quality of the automatically refined narrations. To maintain data quality, we have LLMs focus on one refining step at a time, even though they are capable of handling multiple steps (e.g., dialogue removal and textual correction) in a single response. Additionally, recognizing the importance of contextual clues, such as the co-occurrence of characters and objects in adjacent clips, we implement a batching strategy that allows LLMs to reference surrounding contexts when refining narrations. To further enhance performance, we provide input-output demonstrations for each task, taking advantage of the in-context learning capabilities of LLMs. A manual review of 300 narration samples (see \cref{sec:data-quality}) shows that Movie101v2 exhibits competitive data quality compared to human-refined Movie101.




\begin{figure}[t]
    \begin{center}
        \hspace*{-8pt} 
        \includegraphics[width=0.7\linewidth]{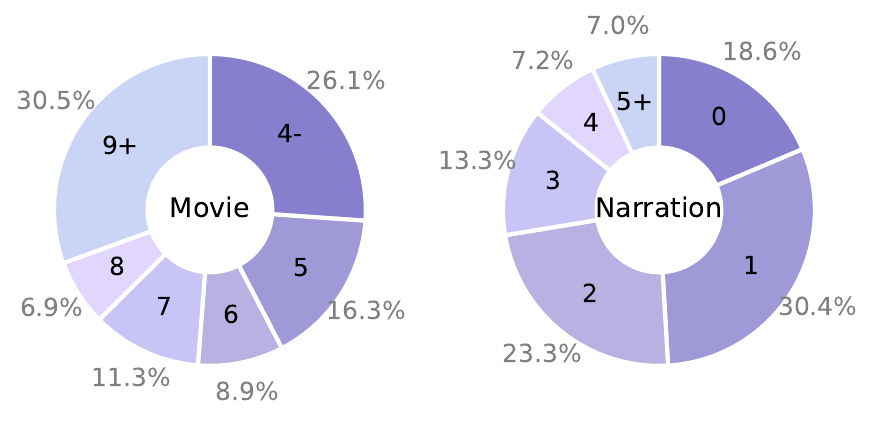}
    \end{center}
    \vspace{-8pt}
\caption{Distribution of character counts in movie casts (left) and narration paragraphs (right).}
\label{fig:role-count}
\vspace{-8pt}
\end{figure}

\begin{figure}[t]
    \begin{center}
        \hspace*{-16pt} 
        \includegraphics[width=0.8\linewidth]{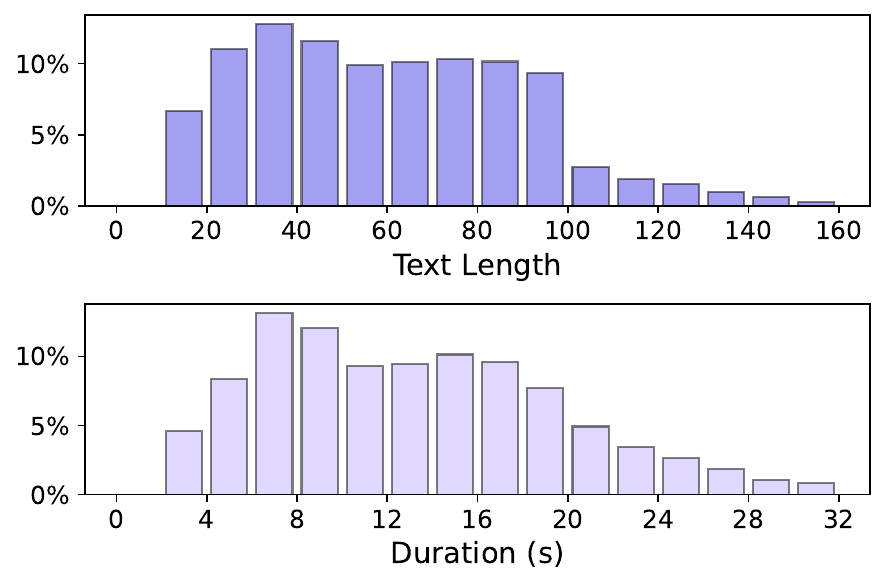}
    \end{center}
    \vspace{-8pt}
\caption{Distribution of narration text length and movie clip duration in Movie101v2-zh.}
\label{fig:length-dist}
\vspace{-8pt}
\end{figure}

\subsection{Data Statistics}

\noindent \textbf{Movies.} Movie101v2 comprises 203 movies totaling 353 hours. On average, each movie features 7.3 characters, with the distribution of character counts detailed in \cref{fig:role-count}. The most popular genres include comedy, romance, and action. The dataset follows the same split as Movie101, with 10 movies each allocated for validation and testing.

\noindent \textbf{Narrations.} We collect 71K video-aligned narration segments and merge them into 46K narration paragraphs, with the length distribution shown in \cref{fig:length-dist}. Detailed statistics comparing Movie101v2 to other datasets are available in \cref{tab:data-stat}. Unlike Movie101, which defines the entire interval between two character dialogues as a single clip for narrating, often resulting in sparse narration annotations over a long clip, Movie101v2 provides denser and more closely video-aligned narrations within corresponding clips. 

\noindent \textbf{Discussion.} By leveraging expert models and LLMs, we establish a streamlined, repeatable, and cost-friendly process for acquiring narration data, laying the groundwork for future dataset expansion. Beyond increasing the data scale, we also enhance the character information to better link narrations with external character knowledge. We hope this enriched data will support the research community in future studies on movie narrating and related tasks, such as temporal grounding in movies~\cite{tsg,movie101}, visual question answering~\cite{moviechat}, and more.
\section{Task: Movie Narration Generation}

This section revisits the task of automatic movie narration generation. We break down the long-term goal of movie narration into three progressive goals (\cref{sec:3.1}), and present a new evaluation framework for narration quality assessment (\cref{sec:3.2}).

\subsection{Task Roadmap}
\label{sec:3.1}

Movie narrations differ from standard video descriptions in that they must not only provide accurate and focused details of visual facts but also infer the plot of the movie by combining information from multiple shots. Effectively narrating a given clip may also rely on the plot's history, as well as cues from sound, character dialogues, and other elements. 
To generate high-quality, deployable movie narrations automatically,
current studies have explored a range of task inputs and outputs. For example, AutoAD highlights the need for various contextual inputs in movie narration generation, using a preceding narration and movie subtitles as additional inputs. 
However, relying on neighbor contexts alone is often insufficient to capture long-term plot history, and depending on previous narrations can lead to cumulative errors. 
Producing context-aware narrations may demand complex and extensive inputs, potentially exceeding the capacity of current models. 
Both Movie101 and AutoAD~II consider the timing of narration generation for practical deployment, with the latter requiring explicit prediction of narration timestamps. However, as our preliminary experiments show, current models struggle with even the most basic narration generation task, suggesting that additional requirements may be unnecessary distractions at this stage.

Given the current state of model development, achieving perfect movie narration remains a challenging, long-term goal. To create a clear roadmap to facilitate developing effective movie narration systems, we propose breaking down this ultimate goal into three progressive stages:


\noindent \textbf{L1. Visual Facts Description.} This stage focuses on providing accurate and comprehensive descriptions of key visual facts in a given movie clip.

\noindent \textbf{L2. Plot Narration.} This stage involves reasoning across multiple shots to describe the plot of the current clip. We emphasize that this stage goes beyond L1, as movies convey plots through the sequence of shots, relying on human cognitive abilities to piece together information fragments into a coherent story. Such a sense of story cannot be achieved by simply listing visual facts from each shot. We provide examples in \cref{sec:l1-l2-diff} to illustrate the distinction between L1 and L2.

\noindent \textbf{L3. Applicable Movie AD.} This final stage aims to produce high-quality narration scripts that are not only accurate and coherent but also appropriately timed and paced, making them directly applicable for practical deployment. 

\noindent The goals of L1 and L2 focus on understanding what occurs within a specific movie clip, i.e., the visual facts and local plots. In contrast, L3 necessitates a more comprehensive understanding, integrating not just individual clips but also the multi-modal context of the entire movie. To achieve the ultimate goal of L3 step by step, we expect first to achieve the foundational goals of L1 and L2, concentrating on understanding and narrating the content of individual clips.

\subsection{Evaluation}
\label{sec:3.2}

\begin{figure}[t]
    \begin{center}
        \hspace*{-6pt}
        \includegraphics[width=0.82\linewidth]{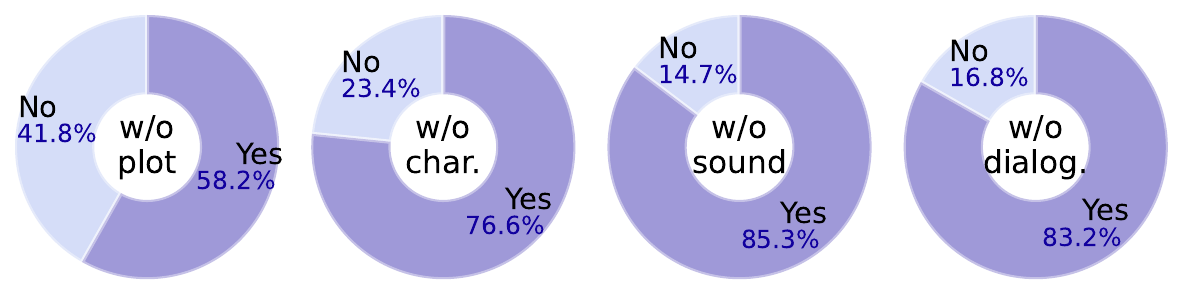}
        \vspace{14pt}
        \hspace*{-16pt}
        \includegraphics[width=0.9\linewidth]{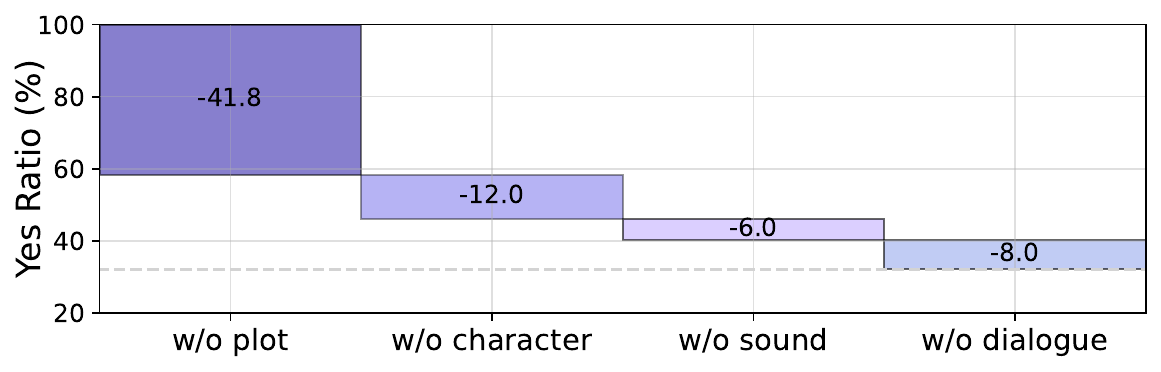}
    \end{center}
    \vspace{-20pt}
\caption{Human evaluation results on the necessity of various multi-modal contextual information for L3 narrations. The top section displays the percentage of "Yes" responses for whether a clip can be accurately narrated without corresponding context. The bottom section shows the "Yes" percentage with all context information being removed progressively.}
\label{fig:context-dependence}
\end{figure}

Existing works primarily evaluate the quality of {the generated narrations by comparing them to the reference ones} (ground truth), using matching-based metrics such as CIDEr~\cite{cider}, BLEU~\cite{bleu}, and ROUGE~\cite{rouge}, feature-based metrics such as BERTScore~\cite{bertscore}, and LLM-based metrics~\cite{autoad-iii}.
However, the reference narrations are derived from deployable ADs created by human experts based on a wealth of contextual information, which is not available to models in the L1 and L2 task settings.
This creates a clear mismatch between {the reference narrations and the task goals}, raising concerns that direct similarity measurements may not provide an appropriate assessment.
To investigate how the ``missing contexts'' affect narration generation, including plot histories, sound, dialogues, and character context (characters are easier to identify {given} prior appearances), we conduct a human evaluation. We randomly sample 1,000 movie clips from the 10 test set movies, and ask annotators whether specific context information is necessary to generate the reference-quality narration.

\begin{figure*}[t]
    \begin{center}
        \includegraphics[width=\linewidth]{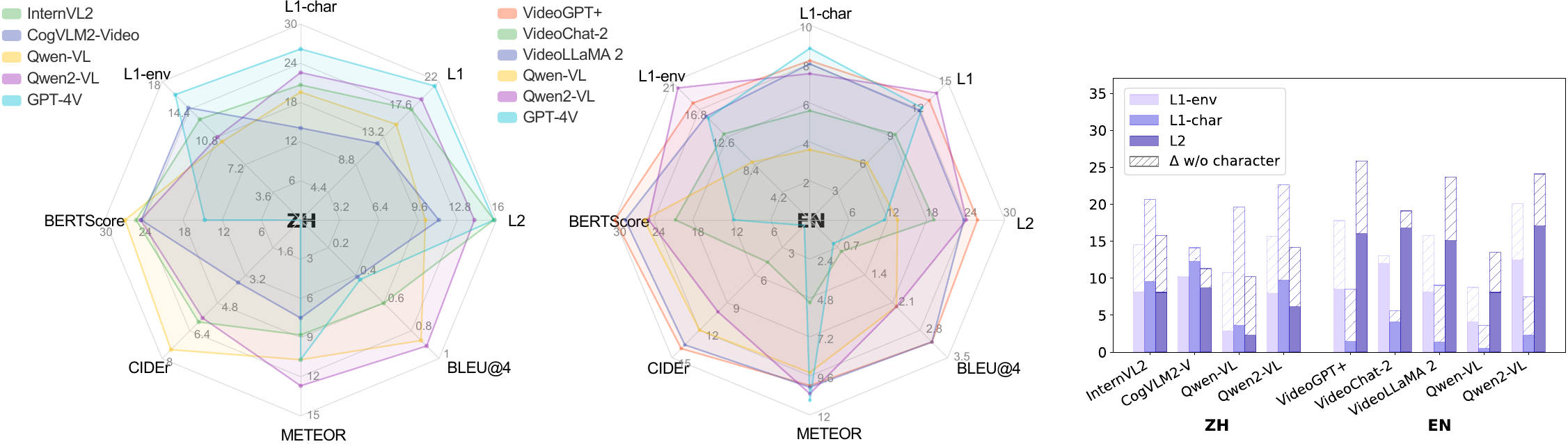}
    \end{center}
    \vspace{-8pt}
\caption{Model performance on Movie101v2. All L1/L2 scores are rescaled to a range of 0-100.}
\label{fig:main-res}
\vspace{-8pt}
\end{figure*}

As shown in \cref{fig:context-dependence} (top), generating reference-quality (or L3) narrations often requires multi-modal context information beyond visual inputs, particularly the plot history and character context. By incrementally removing these contexts (\cref{fig:context-dependence} (bottom)), the ability to generate accurate narrations significantly declines. This highlights the challenge of achieving perfect movie narrations, and supports our argument that reference narrations are often unattainable due to limited model inputs. Consequently, evaluating narrations solely based on their similarity to reference narrations can be misleading and may fail to provide useful feedback for optimizing current narration models. 

To improve the evaluation process, we introduce two new metrics aimed at separately assessing narration quality at the L1 and L2 levels:

\noindent \textbf{L1-Score} evaluates how well the generated narration describes visual facts present in the reference, focusing on (a) \textit{Environment}, including scenes, objects, and events; (b) \textit{Character}, including character names, actions, and emotions.

\noindent \textbf{L2-Score} measures how consistently the generated narration conveys the plot compared to the reference, regardless of the linguistic similarity and specific visual details.

\noindent For each metric, LLM rates on a scale from 0 to 5. These metrics provide a more comprehensive evaluation of model performance,  better aligning with our L1 and L2 task goals. 
\section{Method: Benchmarking and Analysis}

\subsection{Benchmarking}

Large Vision-Language Models (LVLMs) have recently become the leading approach for video-to-text tasks. We benchmark state-of-the-art LVLMs, including VideoGPT+~\cite{videogpt}, VideoChat-2~\cite{videochat-2}, VideoLLaMA~2~\cite{videollama2}, InternVL2~\cite{internvl}, CogVLM2-Video~\cite{cogvlm2}, Qwen-VL~\cite{qwen-vl}, Qwen2-VL~\cite{qwen2-vl}, and GPT-4V, on our movie narration task, which can be broadly categorized into two groups: models that process videos directly and models that process multiple images. The task input consists of a video \(V\) comprising \(L\) frames \(\{f_1, \ldots, f_L\}\), character portraits \(C_{\text{portrait}} = \{p_1, \ldots, p_m\}\) and corresponding character names \(C_{\text{name}} = \{n_1, \ldots, n_m\}\). The goal is to generate a narration \(\hat{y}\) for \(V\). Below, we outline how we adapt the two types of models for this task:

\noindent \textbf{Video Models} typically include a video encoder \(\mathcal{F}(\cdot)\) to extract video features and a projector \(\mathcal{P}(\cdot)\) to convert these features into visual tokens. The LLM then generates narrations based on the visual tokens and task instructions \(I\). To incorporate character information, we treat \(C_{\text{portrait}}\) as additional video frames, and provide \(C_{\text{name}}\) as textual inputs of the LLM. The data flow is defined as:
\[
\scalebox{0.85}{
$\begin{aligned}
    \hat{y} = \mathrm{LLM}\bigg(\mathcal{P}\big(\mathcal{F}(V; C_{\text{portrait}})\big); I; C_{\text{name}}\bigg),
\end{aligned}$
}
\]
where ";" denotes sequence concatenation. We avoid encoding videos and portraits independently with \(\mathcal{P}(\cdot)\), as this can cause the resulting visual tokens 
to lose critical facial information. Instead, we fuse their respective features early, before projection, which has been shown to improve performance in our preliminary experiments.

\noindent \textbf{Multi-image Models} do not explicitly support video input, but can perform inference across multiple video frames. Ideally, we could provide each frame and portrait to the model. However, due to the limited context length (typically supporting up to \(K\) images), we divide the video into \(K\) segments and concatenate adjacent frames into a single image, resulting in \(V'\!=\!\{v'_i\}_{i=1}^K\). Similarly, we concatenate character portraits into \(p'\). To better associate portraits with character names, we also add visual text to the portrait images. The overall process is defined as:
\[
\scalebox{0.85}{
$\begin{aligned}
    \hat{y} = \mathrm{LLM}\bigg(\left\{\mathcal{P}\big(\mathcal{F}(v'_i) \big)| v'_i\in V'\right\}; \mathcal{P}\big(\mathcal{F}(p')\big); I; C_{\text{name}}\bigg).
\end{aligned}$
}
\]
We finetune open-sourced models on the Movie101v2 training set for 3 epochs, freezing the vision encoder and training only the visual projector and LoRA~\cite{lora} adapters of the LLM. For GPT-4V, which cannot be finetuned with our data, we provide a carefully designed task input to encourage in-context learning. This input includes a detailed task description and several randomly retrieved narration examples from the training set to guide the model toward generating narrations in the appropriate style. Implementation details are provided in \cref{sec:imple-model}.

\noindent \textbf{Evaluation.} The L1/L2 scores of Chinese and English narrations are obtained using DeepSeek-V2.5~\cite{deepseekv2} and Llama-3.1-70B-Instruct~\cite{llama3}, respectively. We use these open-sourced models to ensure the best reproducibility, as commercial LLM APIs can be discontinued or updated. Additionally, evaluation results obtained using GPT-3.5-turbo are provided in \cref{sec:gpt-results}. Along with the L1/L2 scores, we also report performance on BLEU, METEOR, CIDEr, and BERTScore for further reference.

\noindent \textbf{Results.} \cref{fig:main-res} shows the performance of various models on Movie101v2. The multi-image model GPT-4V
serves as a strong baseline, particularly in the Chinese setting, despite not being finetuned specifically for this task. Among open-sourced models, VideoGPT+, VideoLLaMA 2, InternVL2, and Qwen2-VL show comparable performance, with some excelling in L1 (visual facts) and others in L2 (plots). Moreover, all models demonstrate improvements when incorporating external character knowledge, as shown in \cref{fig:main-res} (right), highlighting the importance of character understanding in movie narrating. However, despite these gains, all models still exhibit limited task performance, far from being directly applicable, indicating a need for continued research and development. Qualitative results can be found in \cref{sec:supp-qualitative}.


\begin{figure}[t]
    \begin{center}
        \hspace*{0pt}
        \includegraphics[width=0.9\linewidth]{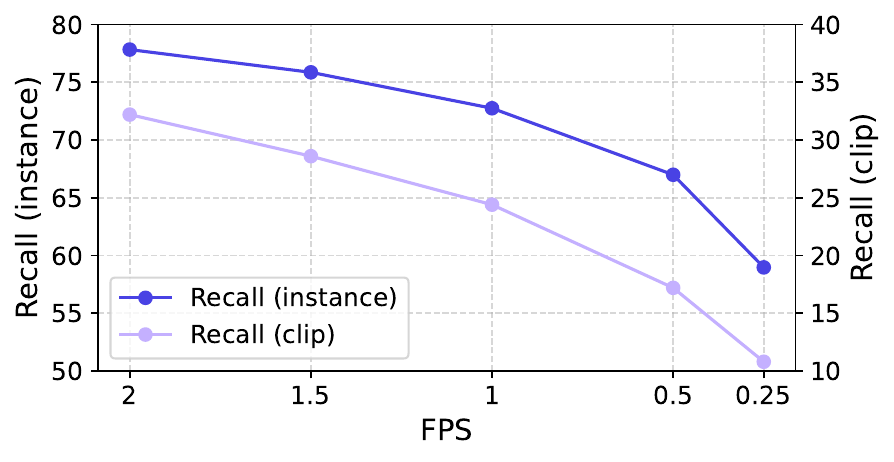}
    \end{center}
    \vspace{-8pt}
\caption{Visual fact recall at varying frame rates. Recall (instance): percentage of recalled visual facts compared to the total visual facts; Recall (clip): percentage of movie clips where all visual facts are recalled.}
\label{fig:recall-fps}
\end{figure}

\subsection{Analysis}
\label{sec:4.2}

Given the limited model performance observed in benchmarking experiments on movie narration, we aim to investigate challenges that models encounter and provide insights for future improvements. We analyze these challenges from the perspectives of visual perception and text generation.

\subsubsection{Visual Perception}

The visual information that a model can perceive from videos depends on (1) its input capacity, i.e., how many frames it can process, and (2) its ability to comprehend the visual inputs. We conduct analytical experiments to pinpoint the essential difficulties related to both aspects, using GPT-4V as a representative for analysis. Given a movie clip \(V\!=\! \{f_i\}_{i=1}^L\) and its ground truth narration \(y\), we first extract the atomic visual facts \(A\!=\!\{a_j\}_{j=1}^N\) mentioned in \(y\) using an LLM (GPT-4). We then query GPT-4V to identify the visual elements in \(A\) that exist in a particular frame \(f_i\) through a binary classification, represented as \(R_i\!=\!\{a_j\}_{j=1}^M, R_i \!\subseteq \!A\). Aggregating \(R_i\) across all frames gives the total visual elements GPT-4V can recall from \(V\). Our analysis covers 500 clips randomly chosen from the Movie101v2 test set, sampled at 2 FPS. We identify several challenges as follows:

\noindent \textbf{Input Capacity Limitation.}
From the 2,954 visual facts in the test narrations, GPT-4V recalls 77.8\% when processing the full 2 FPS input. However, as the frame rate decreases, 
especially below 1 FPS, recall drops sharply (\cref{fig:recall-fps}), indicating a significant information loss due to limited input capacity. Moreover, a human evaluation of 1,000 movie clips sampled at 1 FPS shows that 23.1\% lack the necessary information for accurate narration generation. These results highlight that a limited input capacity represents an early bottleneck for models' visual perception. As most current LVLMs feature a limited visual context length (e.g., 16 frames in VideoChat-2), their ability to perceive long movie clips (e.g., $\geq$ 30 seconds) faces notable challenges. Therefore, improving models to handle longer contexts remains a foremost goal.

\begin{figure}[t]
    \begin{center}
        \hspace*{-10pt} 
        \includegraphics[width=0.9\linewidth]{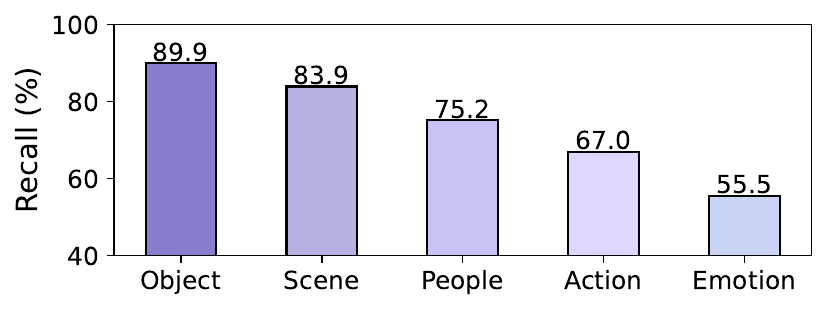}
    \end{center}
    \vspace{-8pt}
\caption{Visual fact recall across various categories.}
\label{fig:recall-category}
\end{figure}

\noindent \textbf{Visual Comprehension Limitation.} As GPT-4V achieves only a 77.8\% recall rate on basic visual recognition tasks, which can further constrain its narration performance, we investigate the specific drawbacks concerning visual comprehension. We divide the visual facts into five categories: objects, scenes, people, actions, and emotions, for a more detailed analysis. As shown in \cref{fig:recall-category}, the model performs better in recognizing objects and scenes but struggles with people-related visual facts, particularly actions and expressions. Since understanding characters and their behaviors is crucial for interpreting movie plots, improving models' comprehension in these aspects deserves further research. 

\begin{table}[t]
\caption{\label{tab:face-recognition}
Results of character face recognition test.
}
\vspace{-4pt}
\centering
\small
\begin{tabular}{@{}l|ccc|c@{}}
\toprule
Model & \# GT & \# Predict. & \# Correct & Precision \\ \midrule
GPT-4V & 1,066 & 1,094 & 477 & 43.6 \\
ArcFace & 1,066 & 713 & 341 & 47.8 \\ \bottomrule
\end{tabular}
\end{table}

\noindent \textbf{Face Recognition Limitation.} Beyond basic visual facts, we also assess the model's ability to identify characters in movie clips based on cast portraits. Following a similar process, given a movie clip and a list of character portraits, we task GPT-4V with identifying the characters appearing in the clip. The ground truth characters are extracted from the reference narration. \cref{tab:face-recognition} shows the results from 500 randomly chosen clips sampled at 1 FPS. While GPT-4V can generally accurately \textit{detect} characters from the video (matching the ground truth in terms of count), it struggles to \textit{identify} who are the specific characters, achieving only 43.6\% precision. It also underperforms when compared to ArcFace~\cite{arcface}, a specialized face recognition model. This limitation in re-identifying cast characters further constrains the model's ability to generate accurate movie narrations. 

\begin{figure}[t]
    \begin{center}
        \includegraphics[width=1\linewidth]{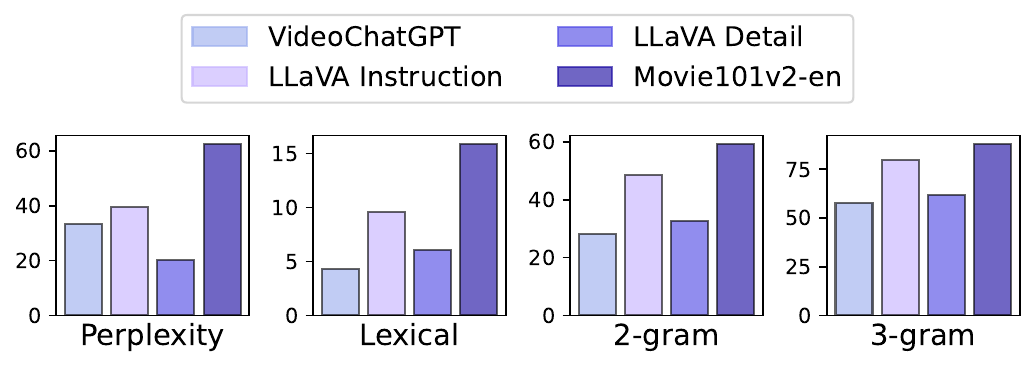}
    \end{center}
    \vspace{-8pt}
\caption{Linguistic properties of different datasets.} 
\label{fig:linguistic-stat}
\vspace{-4pt}
\end{figure}
\begin{figure}[t]
    \begin{center}
        \includegraphics[width=0.85\linewidth]{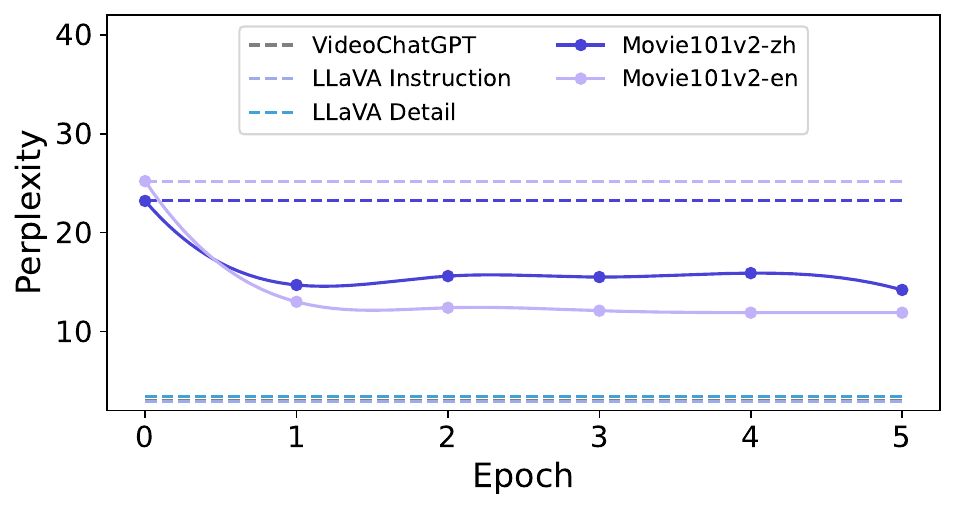}
    \end{center}
    \vspace{-8pt}
\caption{VideoChat-2 perplexity on different datasets. Dotted lines represent the zero-shot perplexity of the model, while solid lines denote the perplexity of the model fine-tuned on Movie101v2.}
\label{fig:ppl-epoch}
\end{figure}

\subsubsection{Textual Generation}

In addition to visual perception, we analyze challenges related to generating textual outputs. From both linguistic and model-fitting perspectives, we observe specific challenges for generating narration texts in Movie101v2. 


\noindent \textbf{Data Linguistics.} We compare the linguistic properties of our movie narration data with those of other datasets commonly used to train LVLMs~\cite{videochatgpt, llava}, using 1,000 samples from each. We evaluate perplexity (a measure of textual fluency and complexity calculated as the exponentiated average negative log-likelihood of a sequence as observed by an evaluator model like GPT-2~\cite{gpt-2}) and n-gram diversity (the ratio of unique n-grams to the total n-gram counts). 
As shown in \cref{fig:linguistic-stat}, Movie101v2 has the highest GPT-2 perplexity and greater lexical diversity (1-gram) and n-gram diversity, indicating more complex narration text compared to other datasets. This complexity presents additional learning challenges for models. 

\noindent \textbf{Model Fitting.} 
We further examine model fitting challenges using perplexity as a measure of how well the model fits the data. 
Lower perplexity indicates a better fit. We demonstrate the perplexity of VideoChat-2 on different datasets in \cref{fig:ppl-epoch}. The model starts with the highest zero-shot perplexity on Movie101v2, suggesting unfamiliarity with the narration text. While perplexity decreases after training, it remains much higher than for other datasets, underscoring the specific challenges in achieving a good fit on complex movie narrations.
\section{Related Works}


Understanding movies by AI has attracted considerable research interest~\cite{graphmovie,movielc,moviellm,moviechat,llama-vid}. MovieNet~\cite{movienet} provides rich annotations of actors and scenes, supporting various proxy tasks like detection, identification, segmentation, and cinematic style prediction. CMD~\cite{cmd} focuses on key scenes coupled with high-level semantic descriptions. YMS~\cite{yms} and SyMoN~\cite{symon} provide plot summaries from movies and TV series for multimodal story comprehension. For a narrative understanding of movies, M-VAD~\cite{m-vad} and MPII-MD~\cite{mpii-md}, combined into LSMDC~\cite{lsmdc}, provide video-aligned movie narrations. PTVD~\cite{ptvd} offers detailed human-written plot descriptions. MAD~\cite{mad} and TVC~\cite{tvr}, originally designed for tasks such as temporal grounding and text-video retrieval, have also demonstrated their value in narration generation. Together, these datasets enable extensive exploration of movie understanding~\cite{autoad,autoad-ii,autoad-iii,movie-understanding-1,movie-understanding-2}. Our work builds upon the recently proposed Movie101~\cite{movie101}, enhancing the data, task definitions, and baseline methods to further support research on generating movie narrations.

\section{Conclusion}


In this work, we present Movie101v2, a large-scale, bilingual dataset designed to advance the development of automatic movie narration generation, making the following contributions: (1) We create a comprehensive dataset using a scalable, repeatable data collection pipeline. (2) We outline a clear task roadmap for achieving the long-term goal of movie narration, and propose corresponding metrics for task evaluation. (3) We benchmark various state-of-the-art models and investigate essential difficulties to motivate future improvements. Our findings reveal a significant gap between the current model capabilities and the requirements for generating applicable movie narrations, underscoring the importance of further research to enable AI-driven solutions that can enhance the movie-watching experience for visually impaired individuals.

\section*{Limitations}

This work introduces an enhanced benchmark for automatic movie narration generation. To accommodate current technological constraints, we have simplified the task of movie narration by focusing on individual movie clips. 
On the one hand, this simplification makes the step-by-step development of deployable movie narration systems more feasible. However, on the other hand, it may constrain more bold and ambitious research aimed at achieving the ultimate goal in a single leap. 


\section*{Ethics Statement}

Movie101v2 is designed to advance research in automatic movie narration generation, with the goal of improving accessibility for visually impaired individuals. 
We address potential ethical concerns as follows:

\noindent \textbf{Data.}
Our dataset consists exclusively of Chinese movies, which may introduce a lack of diversity and potential cultural biases. While most movie narration datasets in current literature are based on English-language movies, we believe that Movie101v2 adds valuable representation to the field. Besides, our dataset contains no personally identifiable information or offensive content. 

\noindent \textbf{Crowdsourcing.} Our data refinement involved a modest amount of crowdsourcing (manual correction of movie cast lists). We compensated annotators with fair wages according to local standards.

\noindent \textbf{Copyrights.}
The movies in our dataset are publicly available on Xigua Video, and our data collection compiles with the service contract of the website\footnote{\url{https://www.ixigua.com/robots.txt}}. To respect copyright, we will release our dataset under highly restrictive permissions, limiting its use strictly to academic research purposes.

\bibliography{custom}

\begin{thebibliography}{45}
\expandafter\ifx\csname natexlab\endcsname\relax\def\natexlab#1{#1}\fi

\bibitem[{Argaw et~al.(2023)Argaw, Lee, Woodson, Kweon, and Heilbron}]{movie-understanding-1}
Dawit~Mureja Argaw, Joon-Young Lee, Markus Woodson, In~So Kweon, and Fabian~Caba Heilbron. 2023.
\newblock Long-range multimodal pretraining for movie understanding.
\newblock In \emph{Proceedings of the IEEE/CVF International Conference on Computer Vision}, pages 13392--13403.

\bibitem[{Bai et~al.(2023)Bai, Bai, Yang, Wang, Tan, Wang, Lin, Zhou, and Zhou}]{qwen-vl}
Jinze Bai, Shuai Bai, Shusheng Yang, Shijie Wang, Sinan Tan, Peng Wang, Junyang Lin, Chang Zhou, and Jingren Zhou. 2023.
\newblock Qwen-vl: A frontier large vision-language model with versatile abilities.
\newblock \emph{arXiv preprint arXiv:2308.12966}.

\bibitem[{Bain et~al.(2020)Bain, Nagrani, Brown, and Zisserman}]{cmd}
Max Bain, Arsha Nagrani, Andrew Brown, and Andrew Zisserman. 2020.
\newblock Condensed movies: Story based retrieval with contextual embeddings.
\newblock In \emph{Proceedings of the Asian Conference on Computer Vision}.

\bibitem[{Chen et~al.(2023{\natexlab{a}})Chen, Ding, Chen, and Jin}]{movielc}
Jieting Chen, Junkai Ding, Wenping Chen, and Qin Jin. 2023{\natexlab{a}}.
\newblock Knowledge enhanced model for live video comment generation.
\newblock In \emph{2023 IEEE International Conference on Multimedia and Expo (ICME)}, pages 2267--2272. IEEE.

\bibitem[{Chen et~al.(2023{\natexlab{b}})Chen, Wu, Wang, Su, Chen, Xing, Zhong, Zhang, Zhu, Lu, Li, Luo, Lu, Qiao, and Dai}]{internvl}
Zhe Chen, Jiannan Wu, Wenhai Wang, Weijie Su, Guo Chen, Sen Xing, Muyan Zhong, Qinglong Zhang, Xizhou Zhu, Lewei Lu, Bin Li, Ping Luo, Tong Lu, Yu~Qiao, and Jifeng Dai. 2023{\natexlab{b}}.
\newblock Internvl: Scaling up vision foundation models and aligning for generic visual-linguistic tasks.
\newblock \emph{arXiv preprint arXiv:2312.14238}.

\bibitem[{Cheng et~al.(2024)Cheng, Leng, Zhang, Xin, Li, Chen, Zhu, Zhang, Luo, Zhao, and Bing}]{videollama2}
Zesen Cheng, Sicong Leng, Hang Zhang, Yifei Xin, Xin Li, Guanzheng Chen, Yongxin Zhu, Wenqi Zhang, Ziyang Luo, Deli Zhao, and Lidong Bing. 2024.
\newblock \href {http://arxiv.org/abs/2406.07476} {Videollama 2: Advancing spatial-temporal modeling and audio understanding in video-llms}.

\bibitem[{DeepSeek-AI(2024)}]{deepseekv2}
DeepSeek-AI. 2024.
\newblock \href {http://arxiv.org/abs/2405.04434} {Deepseek-v2: A strong, economical, and efficient mixture-of-experts language model}.

\bibitem[{Deng et~al.(2019)Deng, Guo, Xue, and Zafeiriou}]{arcface}
Jiankang Deng, Jia Guo, Niannan Xue, and Stefanos Zafeiriou. 2019.
\newblock \href {https://doi.org/10.1109/CVPR.2019.00482} {Arcface: Additive angular margin loss for deep face recognition}.
\newblock In \emph{{IEEE} Conference on Computer Vision and Pattern Recognition, {CVPR} 2019, Long Beach, CA, USA, June 16-20, 2019}, pages 4690--4699. Computer Vision Foundation / {IEEE}.

\bibitem[{Dogan et~al.(2018)Dogan, Li, Sigal, and Gross}]{yms}
Pelin Dogan, Boyang Li, Leonid Sigal, and Markus Gross. 2018.
\newblock A neural multi-sequence alignment technique (neumatch).
\newblock In \emph{Proceedings of the IEEE Conference on Computer Vision and Pattern Recognition}, pages 8749--8758.

\bibitem[{Dubey et~al.(2024)Dubey, Jauhri, Pandey et~al.}]{llama3}
Abhimanyu Dubey, Abhinav Jauhri, Abhinav Pandey, et~al. 2024.
\newblock \href {http://arxiv.org/abs/2407.21783} {The llama 3 herd of models}.

\bibitem[{Gao et~al.(2017)Gao, Sun, Yang, and Nevatia}]{tsg}
Jiyang Gao, Chen Sun, Zhenheng Yang, and Ram Nevatia. 2017.
\newblock \href {https://doi.org/10.1109/ICCV.2017.563} {{TALL:} temporal activity localization via language query}.
\newblock In \emph{{IEEE} International Conference on Computer Vision, {ICCV} 2017, Venice, Italy, October 22-29, 2017}, pages 5277--5285. {IEEE} Computer Society.

\bibitem[{Han et~al.(2023{\natexlab{a}})Han, Bain, Nagrani, Varol, Xie, and Zisserman}]{autoad-ii}
Tengda Han, Max Bain, Arsha Nagrani, Gul Varol, Weidi Xie, and Andrew Zisserman. 2023{\natexlab{a}}.
\newblock Autoad ii: The sequel-who, when, and what in movie audio description.
\newblock In \emph{Proceedings of the IEEE/CVF International Conference on Computer Vision}, pages 13645--13655.

\bibitem[{Han et~al.(2023{\natexlab{b}})Han, Bain, Nagrani, Varol, Xie, and Zisserman}]{autoad}
Tengda Han, Max Bain, Arsha Nagrani, G{\"u}l Varol, Weidi Xie, and Andrew Zisserman. 2023{\natexlab{b}}.
\newblock Autoad: Movie description in context.
\newblock In \emph{Proceedings of the IEEE/CVF Conference on Computer Vision and Pattern Recognition}, pages 18930--18940.

\bibitem[{Han et~al.(2024)Han, Bain, Nagrani, Varol, Xie, and Zisserman}]{autoad-iii}
Tengda Han, Max Bain, Arsha Nagrani, G{\"u}l Varol, Weidi Xie, and Andrew Zisserman. 2024.
\newblock Autoad iii: The prequel-back to the pixels.
\newblock In \emph{Proceedings of the IEEE/CVF Conference on Computer Vision and Pattern Recognition}, pages 18164--18174.

\bibitem[{He et~al.(2016)He, Zhang, Ren, and Sun}]{resnet}
Kaiming He, Xiangyu Zhang, Shaoqing Ren, and Jian Sun. 2016.
\newblock Deep residual learning for image recognition.
\newblock In \emph{Proceedings of the IEEE conference on computer vision and pattern recognition}, pages 770--778.

\bibitem[{Hong et~al.(2024)Hong, Wang, Ding, Yu, Lv, Wang, Cheng, Huang, Ji, Xue et~al.}]{cogvlm2}
Wenyi Hong, Weihan Wang, Ming Ding, Wenmeng Yu, Qingsong Lv, Yan Wang, Yean Cheng, Shiyu Huang, Junhui Ji, Zhao Xue, et~al. 2024.
\newblock Cogvlm2: Visual language models for image and video understanding.
\newblock \emph{arXiv preprint arXiv:2408.16500}.

\bibitem[{Hu et~al.(2022)Hu, Shen, Wallis, Allen{-}Zhu, Li, Wang, Wang, and Chen}]{lora}
Edward~J. Hu, Yelong Shen, Phillip Wallis, Zeyuan Allen{-}Zhu, Yuanzhi Li, Shean Wang, Lu~Wang, and Weizhu Chen. 2022.
\newblock \href {https://openreview.net/forum?id=nZeVKeeFYf9} {Lora: Low-rank adaptation of large language models}.
\newblock In \emph{The Tenth International Conference on Learning Representations, {ICLR} 2022, Virtual Event, April 25-29, 2022}. OpenReview.net.

\bibitem[{Huang et~al.(2020)Huang, Xiong, Rao, Wang, and Lin}]{movienet}
Qingqiu Huang, Yu~Xiong, Anyi Rao, Jiaze Wang, and Dahua Lin. 2020.
\newblock Movienet: A holistic dataset for movie understanding.
\newblock In \emph{Computer Vision--ECCV 2020: 16th European Conference, Glasgow, UK, August 23--28, 2020, Proceedings, Part IV 16}, pages 709--727. Springer.

\bibitem[{Lei et~al.(2020)Lei, Yu, Berg, and Bansal}]{tvr}
Jie Lei, Licheng Yu, Tamara~L Berg, and Mohit Bansal. 2020.
\newblock Tvr: A large-scale dataset for video-subtitle moment retrieval.
\newblock In \emph{Computer Vision--ECCV 2020: 16th European Conference, Glasgow, UK, August 23--28, 2020, Proceedings, Part XXI 16}, pages 447--463. Springer.

\bibitem[{Li et~al.(2023{\natexlab{a}})Li, Peng, Wang, Ge, Liu, Xu, Wang, and Shan}]{ptvd}
Chen Li, Xutan Peng, Teng Wang, Yixiao Ge, Mengyang Liu, Xuyuan Xu, Yexin Wang, and Ying Shan. 2023{\natexlab{a}}.
\newblock Ptvd: A large-scale plot-oriented multimodal dataset based on television dramas.
\newblock \emph{arXiv preprint arXiv:2306.14644}.

\bibitem[{Li et~al.(2023{\natexlab{b}})Li, Wang, He, Li, Wang, Liu, Wang, Xu, Chen, Luo et~al.}]{videochat-2}
Kunchang Li, Yali Wang, Yinan He, Yizhuo Li, Yi~Wang, Yi~Liu, Zun Wang, Jilan Xu, Guo Chen, Ping Luo, et~al. 2023{\natexlab{b}}.
\newblock Mvbench: A comprehensive multi-modal video understanding benchmark.
\newblock \emph{arXiv preprint arXiv:2311.17005}.

\bibitem[{Li et~al.(2023{\natexlab{c}})Li, Wang, and Jia}]{llama-vid}
Yanwei Li, Chengyao Wang, and Jiaya Jia. 2023{\natexlab{c}}.
\newblock Llama-vid: An image is worth 2 tokens in large language models.
\newblock \emph{arXiv preprint arXiv:2311.17043}.

\bibitem[{Lin(2004)}]{rouge}
Chin-Yew Lin. 2004.
\newblock Rouge: A package for automatic evaluation of summaries.
\newblock In \emph{Text summarization branches out}, pages 74--81.

\bibitem[{Liu et~al.(2023)Liu, Li, Wu, and Lee}]{llava}
Haotian Liu, Chunyuan Li, Qingyang Wu, and Yong~Jae Lee. 2023.
\newblock \href {https://arxiv.org/abs/2304.08485} {Visual instruction tuning}.
\newblock \emph{ArXiv preprint}, abs/2304.08485.

\bibitem[{Maaz et~al.(2024)Maaz, Rasheed, Khan, and Khan}]{videogpt}
Muhammad Maaz, Hanoona Rasheed, Salman Khan, and Fahad Khan. 2024.
\newblock \href {http://arxiv.org/abs/2406.09418} {Videogpt+: Integrating image and video encoders for enhanced video understanding}.

\bibitem[{Maaz et~al.(2023)Maaz, Rasheed, Khan, and Khan}]{videochatgpt}
Muhammad Maaz, Hanoona Rasheed, Salman Khan, and Fahad~Shahbaz Khan. 2023.
\newblock Video-chatgpt: Towards detailed video understanding via large vision and language models.
\newblock \emph{arXiv preprint arXiv:2306.05424}.

\bibitem[{OpenAI(2022)}]{whisper}
OpenAI. 2022.
\newblock Introducing whisper.
\newblock \url{https://openai.com/research/whisper}.

\bibitem[{OpenAI(2023)}]{chatgpt}
OpenAI. 2023.
\newblock Introducing chatgpt.
\newblock \url{https://openai.com/blog/chatgpt}.

\bibitem[{PaddleOCR(2022)}]{paddleocr}
PaddleOCR. 2022.
\newblock Paddleocr.
\newblock \url{https://github.com/PaddlePaddle/PaddleOCR}.

\bibitem[{Papineni et~al.(2002)Papineni, Roukos, Ward, and Zhu}]{bleu}
Kishore Papineni, Salim Roukos, Todd Ward, and Wei-Jing Zhu. 2002.
\newblock \href {https://doi.org/10.3115/1073083.1073135} {{B}leu: a method for automatic evaluation of machine translation}.
\newblock In \emph{Proceedings of the 40th Annual Meeting of the Association for Computational Linguistics}, pages 311--318, Philadelphia, Pennsylvania, USA. Association for Computational Linguistics.

\bibitem[{Radford et~al.(2019)Radford, Wu, Child, Luan, Amodei, Sutskever et~al.}]{gpt-2}
Alec Radford, Jeffrey Wu, Rewon Child, David Luan, Dario Amodei, Ilya Sutskever, et~al. 2019.
\newblock Language models are unsupervised multitask learners.
\newblock \emph{OpenAI blog}, 1(8):9.

\bibitem[{Rohrbach et~al.(2015)Rohrbach, Rohrbach, Tandon, and Schiele}]{mpii-md}
Anna Rohrbach, Marcus Rohrbach, Niket Tandon, and Bernt Schiele. 2015.
\newblock \href {https://doi.org/10.1109/CVPR.2015.7298940} {A dataset for movie description}.
\newblock In \emph{{IEEE} Conference on Computer Vision and Pattern Recognition, {CVPR} 2015, Boston, MA, USA, June 7-12, 2015}, pages 3202--3212. {IEEE} Computer Society.

\bibitem[{Rohrbach et~al.(2017)Rohrbach, Torabi, Rohrbach, Tandon, Pal, Larochelle, Courville, and Schiele}]{lsmdc}
Anna Rohrbach, Atousa Torabi, Marcus Rohrbach, Niket Tandon, Christopher Pal, Hugo Larochelle, Aaron Courville, and Bernt Schiele. 2017.
\newblock Movie description.
\newblock \emph{International Journal of Computer Vision}, 123:94--120.

\bibitem[{Soldan et~al.(2022)Soldan, Pardo, Alc{\'a}zar, Caba, Zhao, Giancola, and Ghanem}]{mad}
Mattia Soldan, Alejandro Pardo, Juan~Le{\'o}n Alc{\'a}zar, Fabian Caba, Chen Zhao, Silvio Giancola, and Bernard Ghanem. 2022.
\newblock Mad: A scalable dataset for language grounding in videos from movie audio descriptions.
\newblock In \emph{Proceedings of the IEEE/CVF Conference on Computer Vision and Pattern Recognition}, pages 5026--5035.

\bibitem[{Song et~al.(2023)Song, Chai, Wang, Zhang, Zhou, Wu, Guo, Ye, Lu, Hwang et~al.}]{moviechat}
Enxin Song, Wenhao Chai, Guanhong Wang, Yucheng Zhang, Haoyang Zhou, Feiyang Wu, Xun Guo, Tian Ye, Yan Lu, Jenq-Neng Hwang, et~al. 2023.
\newblock Moviechat: From dense token to sparse memory for long video understanding.
\newblock \emph{arXiv preprint arXiv:2307.16449}.

\bibitem[{Song et~al.(2024)Song, Wang, Sheng, Zhang, Yu, Fan, and Chen}]{moviellm}
Zhende Song, Chenchen Wang, Jiamu Sheng, Chi Zhang, Gang Yu, Jiayuan Fan, and Tao Chen. 2024.
\newblock Moviellm: Enhancing long video understanding with ai-generated movies.
\newblock \emph{arXiv preprint arXiv:2403.01422}.

\bibitem[{Sun et~al.(2022)Sun, Chao, Ji, and Li}]{symon}
Yidan Sun, Qin Chao, Yangfeng Ji, and Boyang Li. 2022.
\newblock Synopses of movie narratives: a video-language dataset for story understanding.
\newblock \emph{arXiv preprint arXiv:2203.05711}.

\bibitem[{Torabi et~al.(2015)Torabi, Pal, Larochelle, and Courville}]{m-vad}
Atousa Torabi, Christopher Pal, Hugo Larochelle, and Aaron Courville. 2015.
\newblock \href {https://arxiv.org/abs/1503.01070} {Using descriptive video services to create a large data source for video annotation research}.
\newblock \emph{ArXiv preprint}, abs/1503.01070.

\bibitem[{Vedantam et~al.(2015)Vedantam, Zitnick, and Parikh}]{cider}
Ramakrishna Vedantam, C.~Lawrence Zitnick, and Devi Parikh. 2015.
\newblock \href {https://doi.org/10.1109/CVPR.2015.7299087} {Cider: Consensus-based image description evaluation}.
\newblock In \emph{{IEEE} Conference on Computer Vision and Pattern Recognition, {CVPR} 2015, Boston, MA, USA, June 7-12, 2015}, pages 4566--4575. {IEEE} Computer Society.

\bibitem[{Wang et~al.(2024)Wang, Bai, Tan, Wang, Fan, Bai, Chen, Liu, Wang, Ge, Fan, Dang, Du, Ren, Men, Liu, Zhou, Zhou, and Lin}]{qwen2-vl}
Peng Wang, Shuai Bai, Sinan Tan, Shijie Wang, Zhihao Fan, Jinze Bai, Keqin Chen, Xuejing Liu, Jialin Wang, Wenbin Ge, Yang Fan, Kai Dang, Mengfei Du, Xuancheng Ren, Rui Men, Dayiheng Liu, Chang Zhou, Jingren Zhou, and Junyang Lin. 2024.
\newblock Qwen2-vl: Enhancing vision-language model's perception of the world at any resolution.
\newblock \emph{arXiv preprint arXiv:2409.12191}.

\bibitem[{Yue et~al.(2023)Yue, Zhang, Hu, Zhang, Wang, and Jin}]{movie101}
Zihao Yue, Qi~Zhang, Anwen Hu, Liang Zhang, Ziheng Wang, and Qin Jin. 2023.
\newblock \href {https://doi.org/10.18653/v1/2023.acl-long.257} {Movie101: A new movie understanding benchmark}.
\newblock In \emph{Proceedings of the 61st Annual Meeting of the Association for Computational Linguistics (Volume 1: Long Papers)}, pages 4669--4684.

\bibitem[{Zhang et~al.(2023)Zhang, Lin, Yang, Wang, Li, Lin, Liu, and Wang}]{movie-understanding-2}
Chaoyi Zhang, Kevin Lin, Zhengyuan Yang, Jianfeng Wang, Linjie Li, Chung-Ching Lin, Zicheng Liu, and Lijuan Wang. 2023.
\newblock Mm-narrator: Narrating long-form videos with multimodal in-context learning.
\newblock \emph{arXiv preprint arXiv:2311.17435}.

\bibitem[{Zhang et~al.(2016)Zhang, Zhang, Li, and Qiao}]{mtcnn}
Kaipeng Zhang, Zhanpeng Zhang, Zhifeng Li, and Yu~Qiao. 2016.
\newblock Joint face detection and alignment using multitask cascaded convolutional networks.
\newblock \emph{IEEE signal processing letters}, 23(10):1499--1503.

\bibitem[{Zhang et~al.(2019)Zhang, Kishore, Wu, Weinberger, and Artzi}]{bertscore}
Tianyi Zhang, Varsha Kishore, Felix Wu, Kilian~Q Weinberger, and Yoav Artzi. 2019.
\newblock Bertscore: Evaluating text generation with bert.
\newblock \emph{arXiv preprint arXiv:1904.09675}.

\bibitem[{Zhu et~al.(2020)Zhu, Song, Dou, Nie, and Zhou}]{graphmovie}
Yutao Zhu, Ruihua Song, Zhicheng Dou, Jian-Yun Nie, and Jin Zhou. 2020.
\newblock \href {https://doi.org/10.18653/v1/2020.acl-main.765} {{S}cript{W}riter: Narrative-guided script generation}.
\newblock In \emph{Proceedings of the 58th Annual Meeting of the Association for Computational Linguistics}, pages 8647--8657, Online. Association for Computational Linguistics.

\end{thebibliography}

\appendix

\section{Data Quality Analysis}
\label{sec:data-quality}

With data quality as our top priority, we have meticulously refined our data pipeline through extensive manual checks. We conduct a quality analysis comparing both Movie101 and our newly collected data, using 300 random samples from each. We first manually check the errors in narration texts, including character dialogue remnants, textual mistakes, and character name mismatches. As shown in \cref{tab:data-quality}, our automatically refined data demonstrates quality comparable to the manually refined Movie101. Additionally, we evaluate the quality of the automatically translated English narrations through human rating (by three annotators) and a back-translation test, where English narrations are translated back into Chinese. Both evaluations, as seen in \cref{tab:data-quality}, confirm the high quality of the automatic translation.

\begin{table}[t]
\caption{\label{tab:data-quality}
Left: Percentage of narrations that contain errors; Right: Human evaluation and back-translation (BT, BLEU@4) results of translated narrations. Human rating 1-5: \textit{terrible}, \textit{poor}, \textit{fair}, \textit{good}, \textit{excellent}.
}
\vspace{-8pt}
\centering
\small
\scalebox{0.92}{
\begin{tabular}{@{}l|cccc|cc@{}}
\toprule
Dataset & Dial. & Text & Name & Avg. & Human & BT \\ \midrule
Movie101 & 0.7 & 3.3 & 2.3 & 2.1 & - & - \\
Movie101v2 & 0.0 & 3.7 & 0.0 & 1.2 & 4.56 & 34.3 \\ \bottomrule
\end{tabular}}
\end{table}
\begin{figure}[t]
    \begin{center}
        \includegraphics[width=1\linewidth]{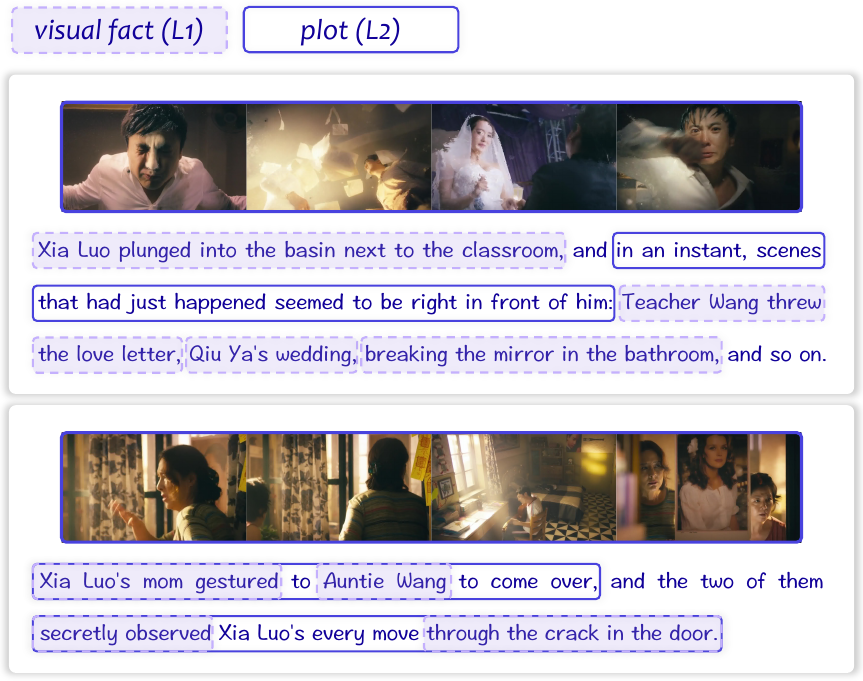}
    \end{center}
    \vspace{-8pt}
\caption{Examples of L1 and L2 narrative elements.}
\label{fig:L1L2-case}
\vspace{-8pt}
\end{figure}

\section{L1 and L2 Differences}
\label{sec:l1-l2-diff}

We illustrate the difference between L1 and L2 narrations in \cref{fig:L1L2-case}. L1 content focuses on basic visual facts that are directly observable in individual frames, while L2 content captures events and plots that are developed by combining L1 elements. 

\section{Implementation Details}

\begin{figure}[t]
    \begin{center}
        \includegraphics[width=\linewidth]{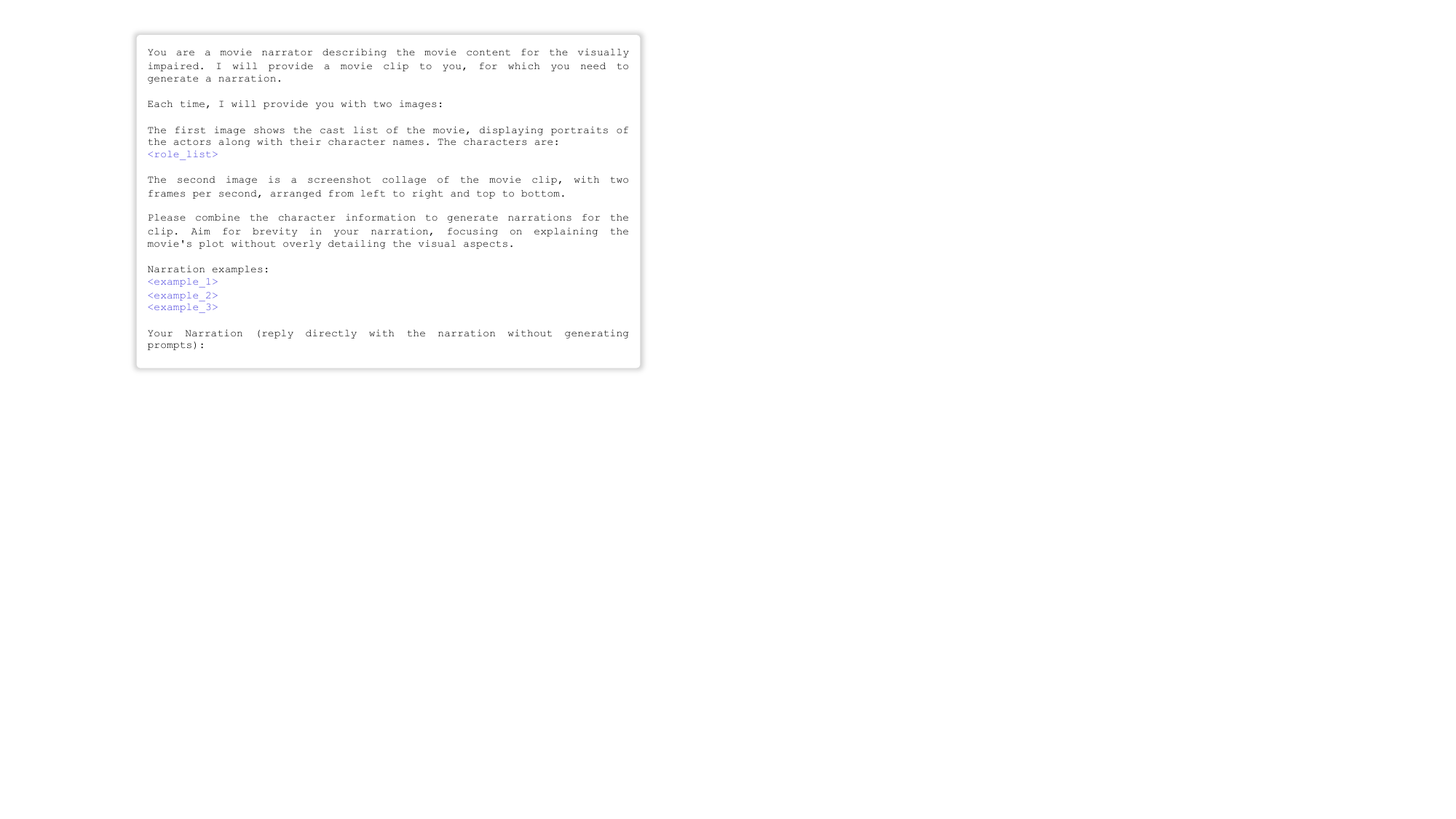}
    \end{center}
\caption{Prompts for GPT-4V to generate narrations.}
\label{fig:prompt-pred}
\end{figure}
\begin{figure}[!ht]
    \begin{center}
        \includegraphics[width=\linewidth]{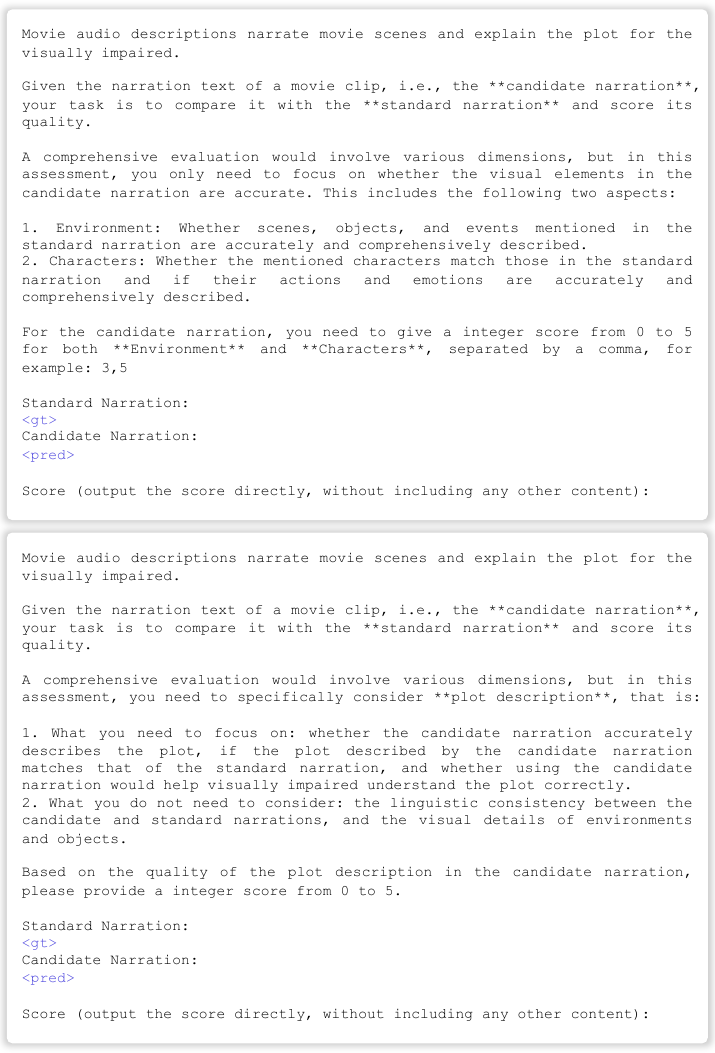}
    \end{center}
\caption{Prompts for producing L1-Score (top) and L2-Score (bottom).}
\label{fig:prompt-eval}
\end{figure}

\subsection{Models}
\label{sec:imple-model}

\noindent \textbf{Video Models.} Models that process video input directly include VideoGPT+, VideoChat-2, VideoLLaMA 2, Qwen2-VL, and InternVL2. For these models, {if not specified,} we set a visual context length of 16 frames, with up to 5 main actor portraits prepended to video frames, taking into account input capacity constraints. In the case of VideoGPT+, which processes video frames into 4-frame chunks, we provide up to 4 actor portraits. {For Qwen2-VL, which accommodates simultaneous video and image input, we provide actor portraits as separate direct image inputs alongside the video.} 
We train the models on the 46.0K narration paragraphs for 3 epochs and evaluate their performance on the 10 test set movies. All other training configurations are consistent with the official documentation\footnote{\scriptsize\url{https://github.com/mbzuai-oryx/VideoGPT-plus}}\footnote{\scriptsize\url{https://github.com/OpenGVLab/Ask-Anything/tree/main/video_chat2}}\footnote{\scriptsize\url{https://github.com/DAMO-NLP-SG/VideoLLaMA2}}\footnote{\scriptsize\url{https://github.com/QwenLM/Qwen2-VL}}\footnote{\scriptsize\url{https://internvl.github.io/blog/2024-07-02-InternVL-2.0/}}.

\noindent \textbf{Multi-image Models.} Models that process multiple images include Qwen-VL and GPT-4V. For Qwen-VL, we keep \(K\!\leq\!4\) segments per video and concatenate the frames sampled at 1 FPS within each segment. Since the model resizes input images to 448px squares, we apply a carefully designed strategy to concatenate both the frames and character portraits while maintaining a balanced aspect ratio. All other training details follow the official configurations for the third training stage of Qwen-VL\footnote{\scriptsize\url{https://github.com/QwenLM/Qwen-VL}}. GPT-4V handles image understanding through configurable high- and low-resolution settings\footnote{\scriptsize\url{https://platform.openai.com/docs/guides/vision}}. In the high-resolution mode, images are scaled to fit within a predetermined size and divided into 512px-square patches, and each patch is represented as a sequence of tokens. For frame images, we apply the high-resolution setting and tailor our frame concatenation strategy to align with the patching approach. We apply the low-resolution setting for portrait images. The API version used is \texttt{gpt-4-0314}. We detail the prompt that guides the model to generate movie narrations in \cref{fig:prompt-pred}. 

\noindent \textbf{ArcFace.} In \cref{sec:4.2}, we evaluate the character face recognition performance of ArcFace. The character faces are first detected and aligned using MTCNN~\cite{mtcnn}, where the minimum face size is set to 30 pixels. We then extract their face features using the ArcFace model with a ResNet-50~\cite{resnet} backbone. Cosine similarities between the face features from frames and portraits are calculated to identify characters detected in the videos.

\begin{table*}[t]
\caption{\label{tab:gpt-res}
Model performance evaluated by GPT-3.5-turbo.
}
\vspace{-4pt}
\centering
\small
\begin{tabular}{@{}c|l|cccccccc@{}}
\toprule
\multirow{2}{*}{} & \multirow{2}{*}{Model} & \multicolumn{4}{c}{w/o character} & \multicolumn{4}{c}{w/ character} \\ \cmidrule(l){3-10} 
 &  & L1-Env & L1-Char & \textbf{L1} & \multicolumn{1}{c|}{\textbf{L2}} & L1-Env & L1-Char & \textbf{L1} & \textbf{L2} \\ \midrule
\multirow{5}{*}{ZH} & InternVL2 & \textbf{21.6} & 34.3 & 27.9 & \multicolumn{1}{c|}{38.1} & 23.9 & 38.7 & 31.3 & \textbf{40.0} \\
 & CogVLM2-video & 21.3 & \textbf{37.0} & \textbf{29.1} & \multicolumn{1}{c|}{36.4} & 22.5 & 34.5 & 28.5 & 37.4 \\
 & Qwen-VL & 17.3 & 30.7 & 24.0 & \multicolumn{1}{c|}{\textbf{39.2}} & 20.7 & 35.1 & 27.9 & 33.6 \\
 & Qwen2-VL & 21.1 & 34.8 & 27.9 & \multicolumn{1}{c|}{35.2} & 24.2 & 39.8 & 32.0 & 37.1 \\
 & GPT-4V & - & - & - & \multicolumn{1}{c|}{-} & \textbf{34.9} & \textbf{46.5} & \textbf{40.7} & 36.4 \\ \midrule
\multirow{6}{*}{EN} & VideoGPT+ & 14.7 & 28.6 & 21.6 & \multicolumn{1}{c|}{\textbf{38.0}} & 18.2 & 34.1 & 26.1 & \textbf{40.3} \\
 & Videochat2 & 6.8 & 23.0 & 14.9 & \multicolumn{1}{c|}{33.3} & 7.6 & 24.5 & 16.0 & 34.1 \\
 & Videollama2 & 13.8 & 28.6 & 21.2 & \multicolumn{1}{c|}{37.2} & 19.1 & 34.0 & 26.5 & 39.2 \\
 & Qwen-VL & 8.6 & 24.7 & 16.7 & \multicolumn{1}{c|}{33.4} & 12.5 & 29.0 & 20.8 & 35.7 \\
 & Qwen2-VL & \textbf{17.3} & \textbf{29.1} & \textbf{23.2} & \multicolumn{1}{c|}{37.2} & 21.7 & 34.8 & 28.2 & 39.2 \\
 & GPT-4V & - & - & - & \multicolumn{1}{c|}{-} & \textbf{27.7} & \textbf{35.0} & \textbf{31.4} & 35.4 \\ \bottomrule
\end{tabular}
\end{table*}

\subsection{Evaluation}

We introduce two evaluation metrics, L1-Score and L2-Score, in \cref{sec:3.2}. The prompts guiding the LLMs to produce these scores are provided in \cref{fig:prompt-eval}. All the LLMs adopt a zero temperature for deterministic decoding.

\section{Additional Results}

\subsection{GPT-3.5 Results}
\label{sec:gpt-results}

In addition to the evaluation results obtained using open-source LLMs in \cref{fig:main-res}, we also provide the results evaluated using GPT-3.5-turbo (\texttt{gpt-3.5-turbo-1106}) in \cref{tab:gpt-res}. These results align closely with those in \cref{fig:main-res}, indicating the consistency of our evaluation framework across different evaluators. However, one exception is that in the English setting, GPT-3.5 shows a clear preference for the output of GPT-4V, which contrasts with Llama-3.1's evaluation.

\subsection{Qualitative Results}
\label{sec:supp-qualitative}

\begin{figure*}[ht]
    \begin{center}
        \includegraphics[width=0.95\linewidth]{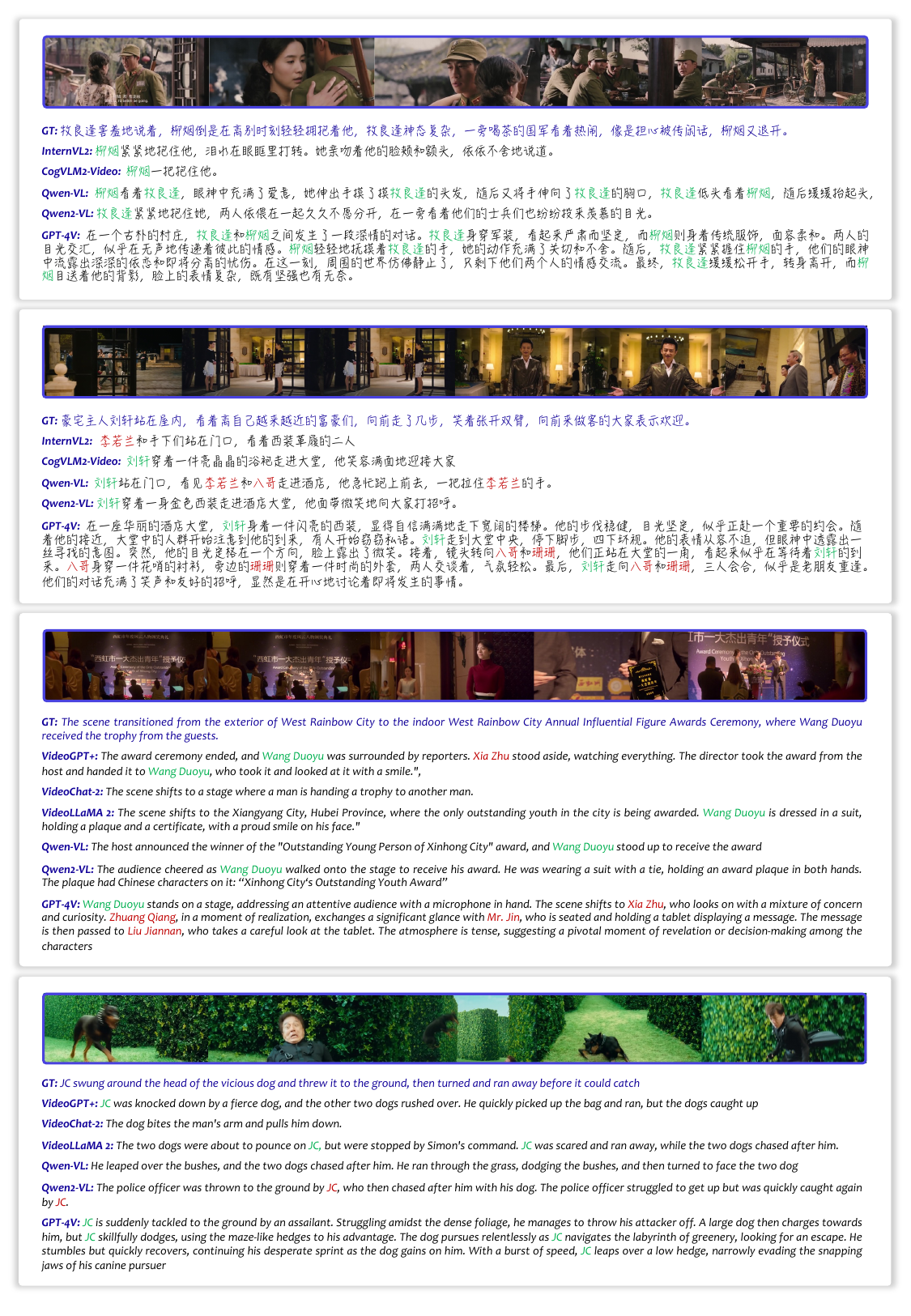}
    \end{center}
    \vspace{-8pt}
\caption{Qualitative results from our baseline models on Movie101v2.}
\label{fig:supp-qualitative}
\vspace{-8pt}
\end{figure*}

We present qualitative results from the baseline models in \cref{fig:supp-qualitative}. The models can often successfully recognize static visual elements, such as scenes, objects, and characters within movie frames. However, they struggle with accurately describing fine-grained events or visual details, such as human interactions or subtle emotions of characters, which often require advanced visual perception and the ability to capture visual changes across consecutive frames. The gap between model-generated narrations and human-created ones underscores the need for further advancements to achieve practical automatic movie narration.
\balance

\end{document}